# The Interaction of Entropy-Based Discretization and Sample Size: An Empirical Study

Casey Bennett[1,2]

[1]Centerstone Research Institute
Nashville, TN, USA
Casey.Bennett@CenterstoneResearch.org

[2]School of Informatics and Computing
Indiana University
Bloomington, IN, USA

## ABSTRACT
An empirical investigation of the interaction of sample size and discretization – in this case the entropy-based method CAIM (Class-Attribute Interdependence Maximization) – was undertaken to evaluate the impact and potential bias introduced into data mining performance metrics due to variation in sample size as it impacts the discretization process. Of particular interest was the effect of discretizing within cross-validation folds averse to outside discretization folds. Previous publications have suggested that discretizing externally can bias performance results; however, a thorough review of the literature found no empirical evidence to support such an assertion. This investigation involved construction of over 117,000 models on seven distinct datasets from the UCI (University of California-Irvine) Machine Learning Library and multiple modeling methods across a variety of configurations of sample size and discretization, with each unique "setup" being independently replicated ten times. The analysis revealed a significant optimistic bias as sample sizes decreased and discretization was employed. The study also revealed that there may be a relationship between the interaction that produces such bias and the numbers and types of predictor attributes, extending the "curse of dimensionality" concept from feature selection into the discretization realm. Directions for further exploration are laid out, as well some general guidelines about the proper application of discretization in light of these results.

**Keywords:** Data Mining; CAIM; Discretization; Sample Size

## 1. INTRODUCTION

Discretization is the process of converting continuous, numeric variables into discrete, nominal variables. In the data mining realm, the discretization process is just one of a number of possible pre-processing steps that may be utilized in any given project. Other pre-processing steps may include feature selection [1], normalization, and class rebalancing (e.g. SMOTE) [2], among others. These data preparation characteristics combine with dataset characteristics – such as sample size, number of attributes, and types of attributes – to create a set of factors that interact to affect the final modeling outcome. These interactions constitute a poorly understood "ecosystem" of factors external to the modeling method itself (e.g. Naïve Bayes, Neural Network). Hand [3] has argued that the effort devoted to understanding this ecosystem is disproportionate to the amount of energy put into developing new methods. Indeed, understanding these interactions and their effects may lead to comparable improvement in modeling outcomes and generalizability of models beyond development of new methods themselves.

Two critical yet empirically unanswered questions around discretization are 1) the impact of variable sample size on the discretization process, particularly entropy-based discretization that relies on patterns in the data itself, and 2) the bias introduced by discretizing within or outside of cross-validation folds during model performance evaluation. This study attempts to empirically evaluate both of these issues. The background for both of these questions is as follows.

Previous research has identified significant interactions between sample size and feature selection affecting the overall accuracy of classifier models produced, as well as the number of features selected [4,5]. These findings are further supported by empirical evidence from applied settings in cancer prognosis prediction [6] and biogeography [7]. This issue has been widely identified in the cancer arena as it relates to the production of "predictive gene lists" (PGL's) for use in diagnosing and treating cancer patients based on clinical and microarray data [6,8,9], as well as genome-wide studies of complex traits in general [10]. In short, smaller sample sizes have been shown to undermine the consistency and replicability of both the reported accuracy and final selected feature sets in these domains [6]. Ein-Dor et al. [6] actually calculated the necessary sample sizes to produce robust, replicable PGL's as being in the thousands, not hundreds as is typically used in many genetic studies. Furthermore, the needed sample size varies depending on the number and types of features analyzed.

The effects of sample size on data mining accuracy, feature selection, and genetic/clinical prediction is thus well established in the literature. However, the relationship between sample size and discretization – particularly entropy-based discretization that relies on the data itself (described below) – is not well established in the literature. Given that entropy-based discretization methods are dependent on the data itself, it should be reasonably suspected that they may be prone to variations in dataset characteristics, e.g. sample size. The question remains as to what impact, if any, disparate sample sizes may have on discretization methods, as well as what bias may be introduced

*Author Contact Info:*
*Casey Bennett*
*Dept. of Informatics*
*Centerstone Research Institute*
*365 South Park Ridge Road*
*Suite 103*
*Bloomington, IN 47401*
*(812)337-2302*
*Casey.Bennett@CenterstoneResearch.org*

when sample sizes are small. To that end, this study focuses on the impact of sample size on one entropy-based discretization method, CAIM (Class-Attribute Interdependence Maximization, defined below), across multiple datasets (exhibiting various dataset characteristics) and classifier methods. All other aspects of modeling (e.g. feature selection) were held constant. This represents a targeted empirical evaluation intended to minimize the number of conflating factors while accounting for potential variability associated with dataset characteristics and/or classifier methods used.

As to the second question, data mining and machine learning literature over the last fifteen years has repeatedly stated that discretizing external to individual cross-validation folds may result in optimistic bias in performance results [11,12,13]. A thorough review of the discretization literature, including conference papers, found that these statements apparently trace back to a paper written by Kohavi and Sahami [14], who state "discretizing all the data once before creating the folds for cross-validation allows the discretization method to have access to the testing data, which is known to result in optimistic error rates." It is important to note that this statement has no data, evidence, or citation associated with it, although it has often been referenced in the literature over the following decade and a half. More interestingly, many papers that claim to compare discretization methods make no explicit mention of how they conduct discretization relative to cross-validation [15,16], nor do many of them include the baseline case of no discretization ([12,14]. Many of them also evaluate only one or two classifier methods [13,14,17,18], which given potential interaction between dataset characteristics and classifier methods, is concerning. As such, empirically investigating such a statement that has become accepted fact would seem an important contribution to the literature [19].

There are many discretization methods in existence. They range from simpler methods such as equal-bins (or equal-widths, choosing a number of equal intervals and dividing the data into each one) and equal-frequency (choosing the bin size by percentage and equally dividing the data into bins by that given frequency, e.g. 25%, 25%, 25%, and 25%) to more complicated methods utilizing the class labels of the target variable to inform the cut-point values of the intervals in the predictor variable. Equal-bins and equal-frequency methods are examples of unsupervised discretization methods, while the latter approaches are considered supervised methods. Examples of supervised methods include chi-squared based methods and entropy-based methods. Chi-squared based methods use chi-squared criterion to establish cut-points by testing the independence of adjacent intervals relative to the class labels. Entropy-based methods are rooted in information theory and measure the minimal amount of information needed to identify the correct label for a given instance [20,21].

CAIM is a form of entropy-based discretization that attempts to maximize the available "information" in the dataset by delineating categories in the predictor variables that relate to classes of the target variable using an iterative approach. CAIM, like all entropy-based methods, works by identifying and using patterns in the data itself in order to improve classifier performance [15]. CAIM has shown promising increases in performance in the literature [15,22,23].

## 2. METHODS

It can be assumed that a larger dataset in the same domain with the same feature set contains more information than a smaller dataset, within bounds (or as a case of diminishing returns). Whereas, given a random discrete variable X with values ranging ($x_i$ ... $x_n$):

$$H(X) = \sum_{i=1}^{n} p(x_i)I(x_i) = -\sum_{i=1}^{n} p(x_i)\log_b p(x_i)$$

represents the principal formula of information theory – where H(X) equals the entropy value of X, p() is the probability distribution, I() is the self-information measure, b is the log base (often 2), and n is the sample size [24]. Within a sample that follows some statistical distribution or observable pattern (the aim of most data mining applications), increasing sample size will refine the probability distribution of the sample as a limit of the function of the true population size. In other words, as the sample size approaches the true population size, the probability distribution of the sample approaches the probability distribution of the actual population. Alternatively, one can conceptualize that increasing sample size "fills in" the probability distribution, mitigating the effects of outliers and reducing the perceived "randomness" that may occur with smaller samples. Random sampling is, of course, intended to ameliorate this precise issue. However, in many domains – particularly the real-world datasets to which data mining is often applied – random sampling is often not possible or of indeterminate degree.

For the second question, we have no theoretical background as to why discretizing within or outside cross-validation folds may bias performance results. In general, the assumption in the data mining realm is that allowing modeling processes (including pre-processing methods such as feature selection) to have access to both the training and test data will optimistically bias results. However, given that discretization processes are significantly dependent on having a *true* distribution from which to work (as explained in the preceding paragraph), there could be some doubt about the effectiveness of discretization when working off of *partial* distributions within each fold. Even when stratified, the stratification is typically based on the target variable to be predicted, rather than the predictor variables to be discretized. In fact, stratification based on the target variable may actually cause the predictor variables to *not* be randomly sampled respective of their distributions. Moreover, in the standard 10-fold cross validation, only one tenth of the actual data is used in each fold to measure performance. The odds that the distribution of each predictor variable in every fold would represent or even approximate the *true* distribution are questionable. If that is the case, then the performance results obtained may be erratic.

**Table 1. Datasets Employed**

| Dataset | # of Instances | # of Attributes | # of Numeric Attributes | Attribute Types |
|---|---|---|---|---|
| Abalone | 4177 | 8 | 7 | Categorical, Numeric |
| Adult | 48842 | 14 | 0 | Categorical, Integer |
| Contraceptive | 1473 | 9 | 0 | Categorical, Integer |
| Gamma | 19020 | 11 | 11 | Numeric |
| Spambase | 4601 | 57 | 55 | Integer, Numeric |
| Wine Quality | 4898 | 12 | 12 | Numeric |
| Yeast | 1484 | 8 | 8 | Numeric |

This investigation involved construction of over 117,000 models on seven distinct datasets containing greater than

1000 instances from the UCI (University of California-Irvine) Machine Learning Library (see Table 1). Multiple modeling methods were applied across a variety of configurations of sample size and discretization, with each unique "setup" being independently replicated ten times in order to produce a sample distribution of performance results. Test/Training datasets were extracted varying in sample size (n=[(50, 100, 200, 400, 700, 1000]) with stratification; any remaining instances were held out as a true "validation set". Five classification methods were employed including Naïve Bayes [20] Multi-layer Perceptron neural networks [20], Random Forests [25], Log Regression, and K-Nearest Neighbors [26]. Additionally, ensembles were built using a combination of these same methods by employing forward selection optimized by AUC (Area Under the Curve) [27]. Voting by Committee was also performed with those same five methods as well, based on maximum probability [28]. In total, seven unique classifier methods were utilized. Modeling was performed using Knime (Version 2.1.1) [29] and WEKA (Waikato Environment for Knowledge Analysis; Version 3.5.6) [20].

Discretization was handled either by pre-sampling (Pre-CAIM, where discretization had access to all data, including the validation set), post-sampling external to cross-validation (Post-CAIM, access to test/training data only), or post-sampling within the cross-validation folds (Within-CAIM, access to training data only). A baseline setup was also performed with no discretization (No-CAIM). Examples can be seen in Figure 1.

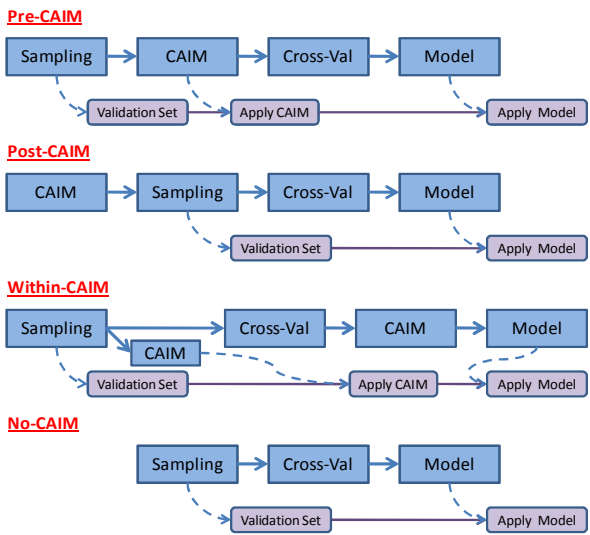

**Figure 1. Discretization Scenarios.**

It is of critical importance to note that for each of the ten replications of the experiment, the whole process was performed from scratch [20]. For instance, for pre-sampling (Pre-Caim) of sample size 400: 1) 400 instances were sampled via stratification, 2) the rest of the samples were held out as an independent validation set, 3) CAIM was applied to the 400-instance sample, 4) classifier models for each of the seven aforementioned methods were constructed on the 400-instance sample and performance measured using 10-fold cross-validation, 5) the CAIM model derived from the 400-instance sample was applied to the validation set, 6) each of the seven classifier models were applied to the validation set and performance measured. These six steps constitute one replication, which was then repeated from the beginning ten times. Each set of ten replications was performed for each combination of sample size (n=50,100,200,400,700,1000) and discretization, including a baseline setup with no discretization.

Performance was then measured via accuracy and AUC (area under curve) via 10-fold cross-validation [20]. Models built using this test/training data were then applied to the validation set to measure actual performance. All models were evaluated using multiple performance metrics, including raw predictive accuracy; variables related to standard ROC (Receiver Operating Characteristic) analysis, AUC, the true positive rate, and the false positive rate [30]; and Hand's H [31]. The data mining methodology and reporting is in keeping with recommended guidelines [3,32], such as the proper construction of cross-validation, testing of multiple methods, and reporting of multiple metrics of performance, among others.

For pre-processing, the target variable in each dataset was re-labeled as "Class" and re-coded to 1 and 0 (the majority class always being 1). It should also be noted that for two datasets – Abalone and Wine Quality – the original target variable was an integer and was thus z-score normalized and converted to a binary variable using a plus/minus mean split. The consequences and assumptions of reduction to a binary classification problem are addressed in Boulesteix et al. [8], noting that the issues of making such assumptions are roughly equivalent to making such assumptions around normal distributions. Additionally, for the Yeast dataset with a multi-class target variable, the target variable was converted to a binary variable as the most common label ("CYT") versus all others. For all continuous predictor variables in each dataset when CAIM was performed, they were first z-score normalized, then discretized via CAIM using the class target variable.

Each of the seven datasets was evaluated across 6 different sampling sizes and seven different modeling methods, across 4 discretization setups (Pre-CAIM, Post-CAIM, Within-CAIM, No-CAIM), equating to 1176 "tests" per dataset. As each "test" was replicated 10 times, the total was 11,760 tests. As the focus here is on evaluating sample size performance, rather than individual dataset performance, an alternative conceptualization is that for each replicate of the 6 sample sizes, there were 147 "tests" per sample size (7 datasets times 7 modeling methods across the 4 discretization setups). There were 10 replicates and 6 sample sizes, equating to a total of 11,760 tests (the same as above). Factoring in the 10-fold cross-validation, there were essentially 117,600 total models constructed during the experimental phase, though in reality this is an underestimate due to the use of ensemble and voting methods.

## 3. RESULTS

The overall results across all methods and datasets are summarized in Figure 2 and Figure 3. Figure 2 shows the pattern of AUC by sample size by the four discretization methods (Pre-CAIM, Post-CAIM, Within-CAIM, and No-CAIM) based on cross-validation performance of the training/test data. Figure 3 shows the performance of the exact same models on the validation set for each of those four discretization methods. The results clearly show an over-optimistic bias in terms of AUC when CAIM is used to discretize a small sample of just a few hundred samples (Pre-CAIM). When either no discretization is used (No-CAIM), or discretization is applied to the entire dataset prior to selecting the smaller sub-sample (Post-CAIM), there is no optimistic bias (at least not due to sample size). In fact, No-CAIM and Post-CAIM follow a very similar pattern across samples sizes. Figure 2 shows that when the models were applied to an independent validation set, the effect of applying CAIM to a small sample size in any scenario is mitigated.

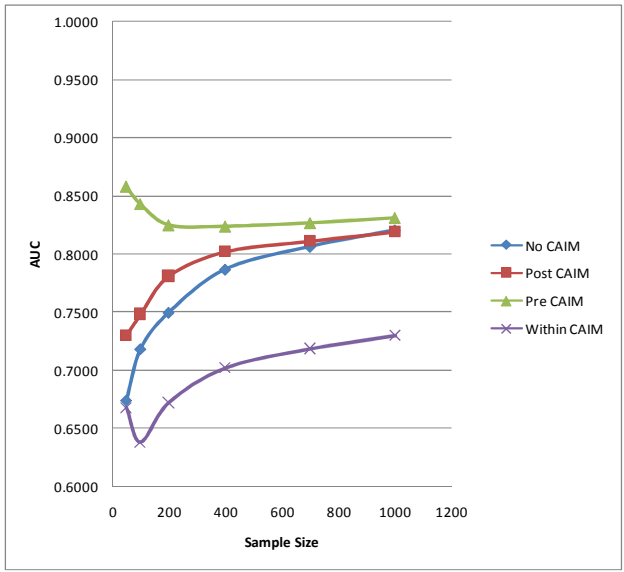

**Figure 2. Cross-Validated Discretization Performance across Sample Size.**

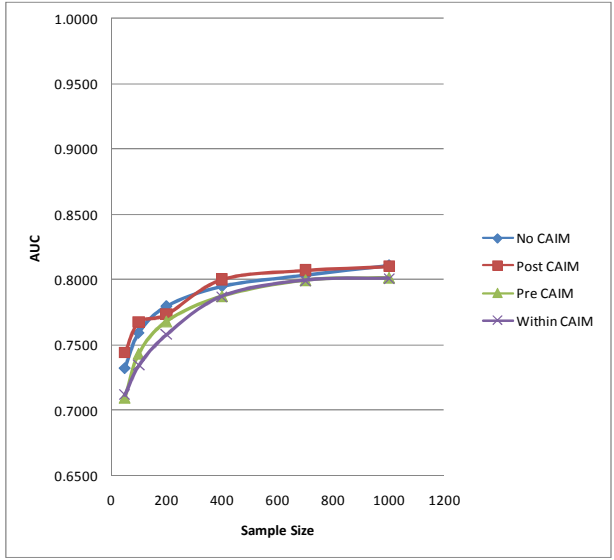

**Figure 3. Validation Set Discretization Performance across Sample Size.**

Additionally, Figures 2 and 3 show the effect of discretizing within cross-validation folds. There is clearly a *negative* bias in terms of performance. When those same models are applied to the validation set, the bias is non-existent. In other words, applying discretization within cross-validation appears to result in incorrect estimates of performance and under-report actual true performance.

Detailed results of cross-validation averaged across all modeling methods and datasets are shown in Table 2.

**Table 2. Cross-Validation Results**

| Sample Size | Avg Accuracy | StdDev Accuracy | Avg AUC | StdDev AUC | Avg H | StdDev H |
|---|---|---|---|---|---|---|
| Pre CAIM | | | | | | |
| 50 | 81.3% | 10.3% | 0.8579 | 0.1122 | 0.571 | 0.237 |
| 100 | 79.5% | 9.1% | 0.8430 | 0.0973 | 0.490 | 0.216 |
| 200 | 77.8% | 8.1% | 0.8248 | 0.0885 | 0.418 | 0.190 |
| 400 | 77.8% | 7.5% | 0.8238 | 0.0811 | 0.396 | 0.175 |
| 700 | 78.0% | 7.2% | 0.8267 | 0.0789 | 0.391 | 0.174 |
| 1000 | 78.5% | 7.3% | 0.8312 | 0.0780 | 0.399 | 0.182 |
| Post CAIM | | | | | | |
| 50 | 71.7% | 10.1% | 0.7300 | 0.1329 | 0.332 | 0.210 |
| 100 | 72.6% | 9.5% | 0.7500 | 0.1157 | 0.321 | 0.202 |
| 200 | 74.3% | 8.0% | 0.7810 | 0.0941 | 0.334 | 0.187 |
| 400 | 76.1% | 7.6% | 0.8019 | 0.0857 | 0.354 | 0.176 |
| 700 | 76.8% | 7.1% | 0.8109 | 0.0802 | 0.360 | 0.176 |
| 1000 | 77.5% | 7.4% | 0.8188 | 0.0812 | 0.374 | 0.183 |
| Within CAIM | | | | | | |
| 50 | 63.0% | 13.5% | 0.6681 | 0.1355 | 0.239 | 0.182 |
| 100 | 59.6% | 15.1% | 0.6384 | 0.1349 | 0.184 | 0.202 |
| 200 | 62.3% | 15.5% | 0.6724 | 0.1313 | 0.199 | 0.204 |
| 400 | 64.4% | 13.7% | 0.7022 | 0.1257 | 0.213 | 0.191 |
| 700 | 68.2% | 12.1% | 0.7187 | 0.1199 | 0.224 | 0.195 |
| 1000 | 68.6% | 11.3% | 0.7300 | 0.1209 | 0.240 | 0.194 |
| No CAIM | | | | | | |
| 50 | 69.7% | 10.4% | 0.7132 | 0.1334 | 0.292 | 0.198 |
| 100 | 72.3% | 8.4% | 0.7493 | 0.1073 | 0.303 | 0.187 |
| 200 | 73.9% | 7.4% | 0.7752 | 0.0942 | 0.317 | 0.175 |
| 400 | 75.1% | 7.4% | 0.7902 | 0.0931 | 0.330 | 0.180 |
| 700 | 75.7% | 7.6% | 0.7995 | 0.0930 | 0.341 | 0.189 |
| 1000 | 76.1% | 7.5% | 0.8059 | 0.0909 | 0.348 | 0.190 |

A separate question is whether CAIM still improves performance over No-CAIM when done externally (e.g. Post-CAIM), as suggested by Kurgan and Cios [15]. Table 2 shows a more detailed view, including standard deviations. Comparing No-CAIM and Post-CAIM based on cross-validation performance seems to suggest some slight improvement (.04 AUC) that diminishes with increasing sample size. However, this slight improvement is diminished to .01 or less across all sample sizes when applied to the validation set. At larger sample sizes (n=700 or greater), the difference between cross-validation and validation set performance is minimal. Given the poor performance of models constructed via CAIM discretization within cross-validation folds, it is unclear whether CAIM discretization actually improves performance or not, and if so under all circumstances.

In terms of the interaction of sample size and CAIM discretization with regards to specific datasets and/or modeling methods, the patterns relative to the various discretization methods were mostly consistent across modeling methods. However, there was more significant variability in the patterns across datasets. The results using AUC can be seen in Tables 3 and 4. Of note, one can observe the consistency in differences between AUC across modeling methods for Pre-CAIM and Post-CAIM at both n=50 and n=1000 in Table 3. These results suggest – at least for the datasets and methods used – that no modeling method is immune to the interaction bias derived from small sample size and CAIM discretization.

**Table 3. Interaction of Discretization and Sample Size – Model Comparison**

| Model | N=50 | | | N=1000 | | |
|---|---|---|---|---|---|---|
| | pre-CAIM | post-CAIM | Diff | pre-CAIM | post-CAIM | Diff |
| Ensemble | 0.8654 | 0.7465 | 0.1189 | 0.8539 | 0.8432 | 0.0108 |
| KNN | 0.8150 | 0.7024 | 0.1126 | 0.7931 | 0.7787 | 0.0144 |
| Log Regression | 0.8328 | 0.6731 | 0.1597 | 0.8405 | 0.8304 | 0.0101 |
| MP Neural Net | 0.8747 | 0.7424 | 0.1323 | 0.8196 | 0.8043 | 0.0153 |
| Naïve Bayes | 0.9003 | 0.7688 | 0.1315 | 0.8476 | 0.8348 | 0.0128 |
| Random Forest | 0.8466 | 0.7262 | 0.1203 | 0.8185 | 0.8074 | 0.0111 |
| Vote | 0.8704 | 0.7504 | 0.1200 | 0.8451 | 0.8331 | 0.0120 |

**Table 4. Interaction of Discretization and Sample Size – Dataset Comparison**

| Dataset | N=50 | | | N=1000 | | |
|---|---|---|---|---|---|---|
| | pre-CAIM | post-CAIM | Diff | pre-CAIM | post-CAIM | Diff |
| Abalone | 0.8003 | 0.7833 | 0.0169 | 0.8286 | 0.8130 | 0.0156 |
| Adult | 0.8317 | 0.7421 | 0.0896 | 0.8700 | 0.8659 | 0.0041 |
| Contraceptive | 0.7055 | 0.6435 | 0.0620 | 0.7168 | 0.7075 | 0.0094 |
| Gamma | 0.9467 | 0.7459 | 0.2007 | 0.8711 | 0.8511 | 0.0200 |
| Spambase | 0.9625 | 0.9122 | 0.0503 | 0.9578 | 0.9576 | 0.0002 |
| Wine Quality | 0.9047 | 0.6682 | 0.2365 | 0.7616 | 0.7871 | -0.0255 |
| Yeast | 0.8537 | 0.6145 | 0.2393 | 0.7869 | 0.7753 | 0.0116 |

Table 4 shows the effect for each individual dataset by AUC. There is significant variability here. However, referencing Table 1, we can observe that the 3 datasets with significantly higher discrepancies in Pre-CAIM and Post-CAIM between n=50 and n=1000 (Gamma, Wine Quality, and Yeast) are all datasets with completely numeric predictors. In other words, all features in that dataset were discretized by CAIM. These 3 datasets all show a roughly .2-.24 discrepancy in AUC at the n=50 sample size. In contrast, datasets with at least one categorical variable (Abalone, Adult, and Contraceptive) range between .02-.08 in terms of AUC at the n=50 level. The remaining dataset (Spambase) is around .05, and though it has some integer variables, it has no categorical variables. It may be that its high overall AUC (~.96 across all instances) limits the range of variation. It's also possible that the reduced number of "unique" values in the integer variables has some impact on the entropy-based discretization process. However, neither of those possibilities can be confirmed nor refuted based on only one dataset. Also, of note, across the 3 datasets containing categorical variables, there was no clear relationship between the number of categorical variables and the impact of CAIM on the discrepancy in AUC between n=50 and n=1000. Again, with only 3 datasets, it is difficult to draw firm conclusions, but this may present an opportunity for further empirical examination.

An additional question is whether the impact of discretizing within or outside of cross-validation folds varies by dataset and/or modeling method, including across varying sample sized. Results are shown in Tables 5 and 6, comparing Within-CAIM and No-CAIM at both n=50 and n=1000.

**Table 5. Effect of Discretizing Within Cross-Validation Folds – Model Comparison**

| Model | N=50 | | | N=1000 | | |
|---|---|---|---|---|---|---|
| | Within-CAIM | No-CAIM | Diff | Within-CAIM | No-CAIM | Diff |
| Ensemble | 0.6791 | 0.7306 | -0.0515 | 0.7895 | 0.8429 | -0.0535 |
| KNN | 0.6828 | 0.6381 | 0.0447 | 0.7430 | 0.7034 | 0.0396 |
| Log Regression | 0.6524 | 0.6933 | -0.0409 | 0.6263 | 0.8206 | -0.1943 |
| MP Neural Net | 0.7005 | 0.7208 | -0.0203 | 0.7440 | 0.8213 | -0.0773 |
| Naïve Bayes | 0.7008 | 0.7418 | -0.0410 | 0.7706 | 0.7993 | -0.0287 |
| Random Forest | 0.6541 | 0.7274 | -0.0733 | 0.7565 | 0.8302 | -0.0737 |
| Vote | 0.6073 | 0.7405 | -0.1332 | 0.6801 | 0.8232 | -0.1431 |

**Table 6. Effect of Discretizing Within Cross-Validation Folds – Dataset Comparison**

| Dataset | N=50 | | | N=1000 | | |
|---|---|---|---|---|---|---|
| | Within-CAIM | No-CAIM | Diff | Within-CAIM | No-CAIM | Diff |
| Abalone | 0.7451 | 0.7771 | -0.0320 | 0.8278 | 0.8301 | -0.0024 |
| Adult | 0.7734 | 0.7343 | 0.0392 | 0.7973 | 0.8376 | -0.0403 |
| Contraceptive | 0.6089 | 0.5756 | 0.0333 | 0.6791 | 0.6863 | -0.0072 |
| Gamma | 0.5310 | 0.7349 | -0.2040 | 0.7080 | 0.8317 | -0.1237 |
| Spambase | 0.8213 | 0.8568 | -0.0355 | 0.8241 | 0.9427 | -0.1186 |
| Wine Quality | 0.6511 | 0.6516 | -0.0005 | 0.6843 | 0.7737 | -0.0894 |
| Yeast | 0.5461 | 0.6622 | -0.1161 | 0.5893 | 0.7388 | -0.1494 |

As can be seen, there is significant variability across datasets and modeling methods. Bias in modeling methods ranges from .04 to -.14 AUC. For the most, it is negative, except for KNN. The bias is fairly consistent across sample sizes, except for Log Regression and MP Neural Networks. Voting based on maximum probability by far suffers the greatest degradation in performance. Across dataset, the impact is highly variable, ranging from .03 to -.2 AUC. The datasets with the greatest impact (Gamma and Yeast) are both datasets with completely numeric predictors. However, the other dataset with completely numeric predictors (Wine Quality) had virtually no difference in performance between Within-CAIM and No-CAIM at n=50, and only a moderate difference at n=1000. The results shown in Table 5 and 6 appear to be highly erratic. This unpredictability complicates any consistent interpretation.

## 4. DISCUSSION

An empirical investigation of the interaction of sample size and discretization methods was performed, utilizing a replicate-based study to produce a sample distribution [20]. This analysis revealed a significant impact of sample size on discretization, in particular the entropy-based method CAIM [15], leading to optimistic bias in performance metrics at lower sample sizes. Previous research has revealed the interaction of sample size with data mining accuracy, feature selection, and genetic/clinical prediction [1,5,6], but the interaction with discretization methods is equally important. Without careful consideration of this interaction, researchers may obtain incorrect performance metrics for constructed models. It is thus important to be cognizant of this factor when considering modeling and study design.

Additionally, the results revealed a significant negative bias when discretization was performed within cross-validation folds relative to the validation set. In other words, performing discretization within cross-validation folds appears to risk under-reporting performance results. This runs contrary to previous assertions in the literature (for which no empirical evidence has been provided or published), that discretizing external to cross-validation folds optimistically biases performance [14]. When examined across modeling methods and datasets, the pattern of bias was highly erratic. The lack of a predictable pattern of impact complicates an interpretation of how such an impact may affect a given data mining experiment and its results.

This interaction between sample size and discretization was consistent across modeling methods – i.e. no classifier was immune. Datasets presented more variability depending on the types (categorical, binary, continuous) and numbers of predictor attributes. However, the precise nature of the interaction between specific types and numbers of attributes and its effect on CAIM and/or discretization in general needs closer examination across a larger number of datasets and/or discretization methods. In the context of this study, it can be determined that such an interaction exists, but its behavior relative to the characteristics of specific datasets remains to be explicitly defined.

There are numerous limitations to this study, many of which demand further research. Only seven datasets were used, and only one specific discretization method was evaluated (results may or may not generalize to other discretization methods, even other entropy-based ones). For the former, this was due to constraints being placed on the datasets (e.g. n>1000) in order to vary sample size and still have a validation set. For the latter, this was due to the experimental intent to hold as many aspects of the study constant while varying other aspects such as sample size, discretization timing within the workflow, and modeling method. For the same reason, feature selection was also not employed, but there may be some yet understood interaction between discretization, feature selection, and/or other dataset characteristics (e.g. sample size) or modeling processes (classifier method employed).

In short, until we understand the interaction of the different components of the modeling ecosystem, any attempts to "compare" different aspects of individual steps may or may not generalize to broader scenarios (such as when different classifier methods are employed). This study is a step, albeit a small one, in that direction. Additional discretization methods must be evaluated. Moreover, the individual datasets selected from UCI Machine Learning Library may or may not be representative of the entire universe of possible datasets. Re-analyzing these results, either in part or in whole, with other datasets may be informative, although computational time and complexity will limit the reach of any one study.

## 5. ACKNOWLEDGMENTS

This project was funded by a grant through the Ayers Foundation and the Joe C. Davis Foundation. The funder had no role in design, implementation, or analysis of this research. The authors would also like to recognize various Centerstone staff for their contributions and support of this effort: Dr. Tom Doub, Dr. April Bragg, Christina Van Regenmorter, and others.